%% file: paper.tex
\documentclass{article}

\usepackage{microtype}
\usepackage{graphicx}
\usepackage{subfigure}

\usepackage{booktabs} 
\usepackage{mathptmx}
\usepackage{color}
\usepackage{amsfonts}       
\usepackage{nicefrac}       
\usepackage{microtype}      
\usepackage{amsmath,graphicx}

\usepackage{amsmath}

\usepackage{outlines}

\usepackage{hyperref}

\usepackage[accepted]{icml2020}

\usepackage{enumitem}

\graphicspath{{../figs/}{./figs}}

\icmltitlerunning{Actionable Attribution Maps for Scientific Machine Learning}

\begin{document}

\twocolumn[
\icmltitle{Actionable Attribution Maps for Scientific Machine Learning}




\begin{icmlauthorlist}
\icmlauthor{Shusen Liu}{casc}
\icmlauthor{Bhavya Kailkhura}{casc}
\icmlauthor{Jize Zhang}{casc}
\icmlauthor{Anna M. Hiszpanski}{pls}
\icmlauthor{Emily Robertson}{pls}
\icmlauthor{Donald Loveland}{pls}
\icmlauthor{T. Yong-Jin Han}{pls}
\end{icmlauthorlist}

\icmlaffiliation{casc}{Center for Applied Scientific Computing, Computing Directorate, Lawrence Livermore National Laboratory, US}
\icmlaffiliation{pls}{Materials Science Division, Physical and Life Science Directorate, Lawrence Livermore National Laboratory, US}


\icmlcorrespondingauthor{Shusen Liu}{liu42@llnl.gov}
\icmlcorrespondingauthor{T. Yong-Jin Han}{han5@llnl.gov}

\icmlkeywords{Explainable AI, Scientific ML, Material Science, Concept Attribution}

\vskip 0.3in
]



\printAffiliationsAndNotice{}

\begin{abstract}

The scientific community has been increasingly interested in harnessing the power of deep learning to solve various domain challenges. However, despite the effectiveness in building predictive models, fundamental challenges exist in extracting actionable knowledge from the deep neural network due to their opaque nature. In this work, we introduce techniques for exploring the behavior of deep learning models by injecting domain-specific actionable concepts as tunable ``knobs'' in the analysis pipeline. By incorporating the domain knowledge, we are not only able to better evaluate the behavior of these black-box models, but also provide scientists with actionable insights that can potentially lead to fundamental discoveries.

\end{abstract}

\input{intro.tex}
\input{related.tex}

\input{method.tex}
\input{results.tex}

\section{Conclusion}
In this work, we introduced a general technique for inferring domain insights from a given predictive model by understanding and manipulating the meaningful variation in the model's input (e.g., SEM images). The ability to turn these explainable ``knobs'' allows us to obtain an actionable understanding of how the prediction is affected by key domain concepts (e.g., porosity, dispersity, size of the material crystal) in the analysis pipeline.
To better understand the combined effects of multiple concepts, we introduced an optimization that allows the model to reveal what attribute combinations will yield a more desirable output.
Since our ability to meaningfully modify and synthesize new SEM images is driven by image editing GANs (e.g.,~\cite{attGAN}), one particular challenge originates from the potential distribution shift from the original image to the reconstructed images (when we synthesize new image tiles using the attributes associate the corresponding lot). Even though the human often can not discern any noticeable difference between the original images and reconstruct ones, these unnoticeable changes can lead to minor prediction shift from the result from the original.


%
%


\section*{Acknowledgements}
This work was performed under the auspices of the U.S. Department of Energy by Lawrence Livermore National Laboratory under Contract DE-AC52-07NA27344. This work is reviewed and released under LLNL-JRNL-811201.

\bibliography{refs.bib}
\bibliographystyle{icml2020}

%
%
%

\end{document}

%% file: intro.tex
\section{Introduction}

Due to the tremendous success of deep learning in commercial applications, there are significant efforts to exploit these tools to solve various scientific challenges.
Unfortunately, these complex models are often considered as black boxes~\cite{holm2019defense} and are extremely hard to interpret.
Besides the inherent model complexity,
scientific data often requires domain knowledge to be understood and to be annotated, which often leads to label sparsity.
Furthermore, instead of focusing on predictive performance, in scientific applications, we particularly value the insights distilled from the model that can potentially advance scientific understanding.
Many existing scientific applications of deep learning focus on building a predictive model for certain experiment output modality (e.g., building a model for predicting the material peak stress given a SEM image).
However, despite their effectiveness in predicting the quantity of interest, we do not have a viable way to evaluate and reason about their decisions to the domain scientists.

\begin{figure}[htbp]
\centering
  \includegraphics[width=1.0\linewidth]{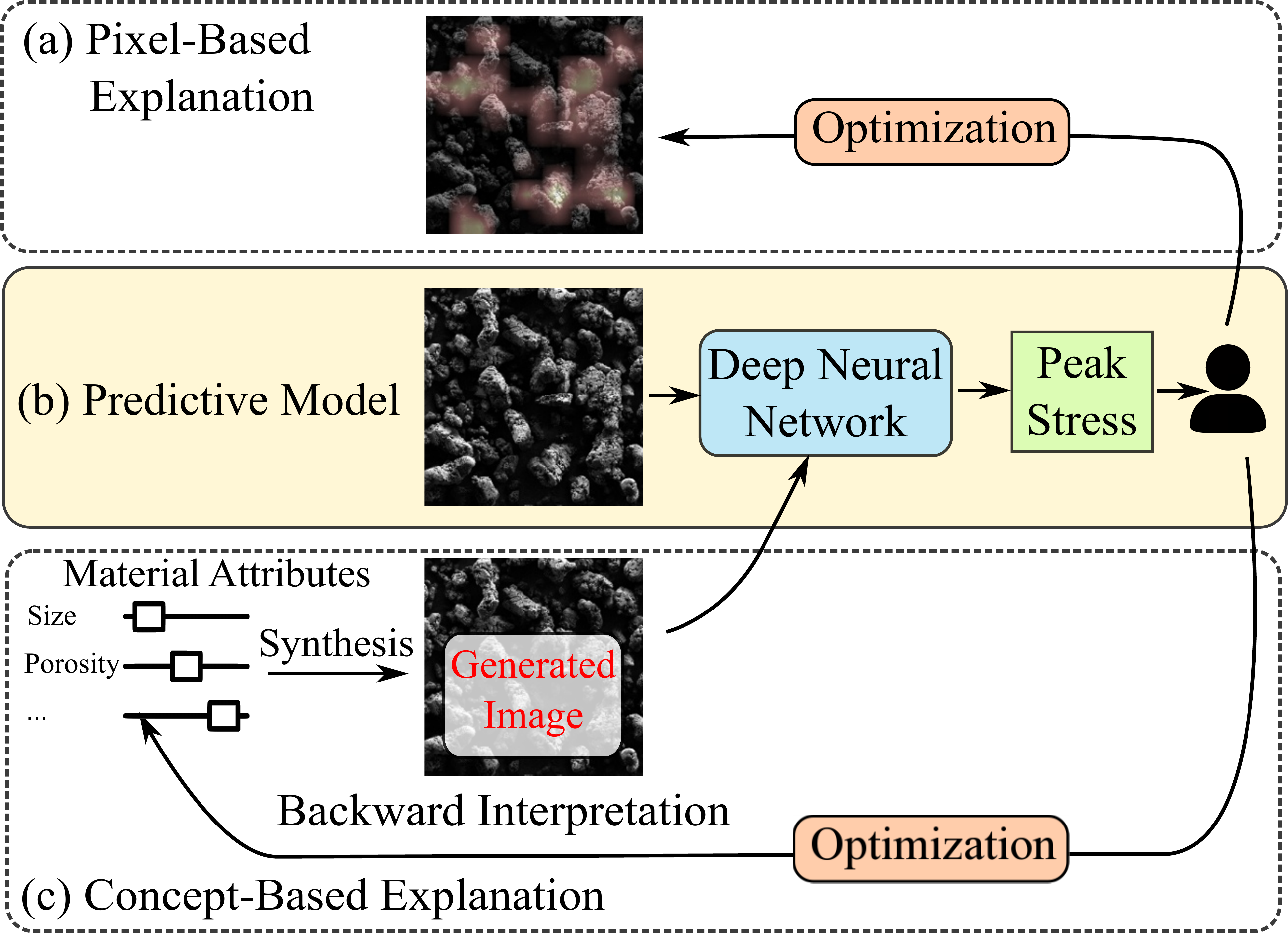}
  \vspace{-3mm}
 \caption{
Overview of the concept-based explanation pipeline. We have a deep neural network model (b) for predicting material peak stress from SEM image.  Instead of trying to attribute the decision to the input pixel space (e.g., GradCAM~\cite{selvaraju2017grad}) (a), we can provide more effective reasoning of the model behavior by injecting domain concepts in the analysis pipeline (c), and explain the behavior using meaningful language (material attributes) of the problem domain.
 }
\label{fig:pipeline}
\end{figure}

One key reason that leads to these challenges is our inability to reason about domain concepts that are meaningful to the scientists in the deep learning pipeline.
As illustrated in Figure~\ref{fig:pipeline}(b), we have a deep learning model that predicts the peak stress of the material given a scanning electron microscope (SEM) image as an input.
The traditional pixel-based attribution (saliency) explanation approaches~\cite{ZeilerFergus2014, bach2015pixel,selvaraju2017grad} for the convolutional neural network (CNN) produces a heat-map (on a per-pixel level) to highlight the region in the image that contributed the most to the prediction (Figure~\ref{fig:pipeline} (a)). Such an approach may work well for natural images, e.g., highlighting the head of the cat when predicting a cat image.
However, this pixel-based attribution is not particularly insightful when trying to explain why certain material has higher peak stress by highlighting pixel in the image as illustrated in Figure~\ref{fig:pipeline}(a). The reason being that we do not necessarily have the answer ourselves and the interpretation in the image pixel space does not correspond to any meaningful or understandable material science concept. Furthermore, a material scientist may be more interested in understanding the effect of only a subset of all possible implicit concepts (e.g., SEM image) which are explicit and are actionable (e.g., particle size in the SEM image).

In this work, we aim to address this fundamental explainability challenge by injecting meaningful concepts post-hoc into the prediction pipeline by utilizing neural image synthesis.
As illustrated in Figure~\ref{fig:pipeline}, we first build a neural image synthesis model that can produce ``fake'' SEM images compliant to user-controlled attributes, e.g., an image with larger or smaller particle size, more porous or less porous material. We then leverage these attributes as the interpretable handles to more effectively reason about the predictive model behavior.
Instead of explaining the model behavior by highlighting the input image, we can now directly answer the questions in the language that the domain scientists understand, i.e., \textit{how does the particle size (or porosity, etc.) impact the peak stress prediction? or what material attributes should be altered to obtain a material with higher peak stress?}
Moreover, compared to the correlation analysis between material attributes and prediction outputs, the proposed method not only produce a per-instance explanation but also generates the corresponding synthesized SEM image that reflects the manifestation of the changes indicated by attribution values. Such images of hypothetical lots can be particularly helpful to the material scientist for gaining intuitive understanding of the synthesis process and potentially revealing other previously unknown variations in the image that is not captured by known attributes.


Our key contributions are listed as follows:
\begin{itemize}[noitemsep,topsep=0pt,parsep=0pt,partopsep=0pt]
  \item Introduce a novel concept-driven reasoning framework for explaining a complex predictive model and showcase its capability in helping domain scientists to obtain actionable insights in feedstock material synthesis;
  \item Demonstrate the effectiveness of conditional neural image synthesis system for capturing the association between domain concepts and intricate image features with extremely small amount of supervised information (in our case, a total of 30 unique labels are used);
\end{itemize}

%
%

%% file: related.tex
\section{Related Works}


The opaque nature of deep neural networks has prompted many efforts for their interpretation.
One key strategy for explaination is attributing the prediction into the model's input domain, most notably for the convolution neural network (CNN).
Various approaches~\cite{SimonyanVedaldiZisserman2013, ZeilerFergus2014, YosinskiCluneNguyen2015, bach2015pixel, lapuschkin2019unmasking} have been proposed to highlight the important region in the image that contributes most to the decision.
We can also approach the attribution explaination scheme from a model agnostics perspective~\cite{RibeiroSinghGuestrin2016, KrausePererNg2016, LundbergLee2017}, e.g., the LIME~\cite{RibeiroSinghGuestrin2016} explains a prediction by fitting a localized linear model for approximating the classification boundary for a given prediction.
However, these attribution is only useful if the input domain is highly interpretable or meaningful to the observer (e.g., natural images), which is not the case in many scientific applications (e.g., SEM images).
Moreover, the assignment of the importance to input is limited in the sense that it can only provide passive correlative information.
To address these challenges counterfactual explaination approaches~\cite{kusner2017counterfactual,narendra2018explaining, Goyal2019, anne2018grounding} have been proposed, e.g., the counterfactual visual explanation~\cite{Goyal2019} work introduces a patch-based image editing and optimization scheme for obtaining interpretable changes in the image domain for altering a prediction.
However, the patch-based editing can severely limit the expressiveness of input the image variation.
In this work, we leverage the concept of the generative counterfactual reasoning \cite{liu2019generative} to faciliate the explanantion of a material science deep learning system by providing a generative model that able to meaningfully edit the model input (e.g., image) via domain-specific attributes.

%% file: method.tex
\section{Method}
\label{sec:method}


\subsection{Domain-Guided Image Synthesis}
\label{sec:imageSynthesis}

We aim to accelerate the material development process by leveraging the modeling capability of deep learning and understand the relationship between the salient feature of observed data (e.g., SEM images) and the material's characteristics.
Specifically, we focus on feedstock material (i.e., raw material for more complex products), where the compressive strength-related properties are crucial.
The experiment involves 30 different batches of material samples (referred to as lots).
Each lot is analyzed with a Scanning Electron Microscope (SEM), where several high-resolution scans are produced. To aid the learning process, the high-resolution scanned image is then divided into multiple smaller tiles ($1000\times1000$ pixels), which lead to more than 60k greyscale images across 30 lots.

To help understand material appearance in each lot, the material scientists coarsely estimate the following properties -- \emph{size}, \emph{porosity}, \emph{polydispersity}, \emph{facetness}, by examining several images per-lot and average the estimates from multiple experts.
The following are the meaning of each property and specific features in the image the scientists are looking after:
\emph{size} -- the average size of crystals;
\emph{porosity} -- how ``holey'' the crystals are, i.e., does it look like they have a lot of small pin-prick holes on the surface or are they solid;
\emph{dispersity} -- how varied the size of the particles are, i.e., how broad is the size distribution;
\emph{facetness} -- do the crystals look rounded/smooth at edges or do they have flat faces that meet at different angles to give a faceted structure.
As a result, a total of only 30 labels are captured, which can be considered as extremely small for most deep learning tasks.

\begin{figure}[!htbp]
\centering
  \includegraphics[width=0.99\linewidth]{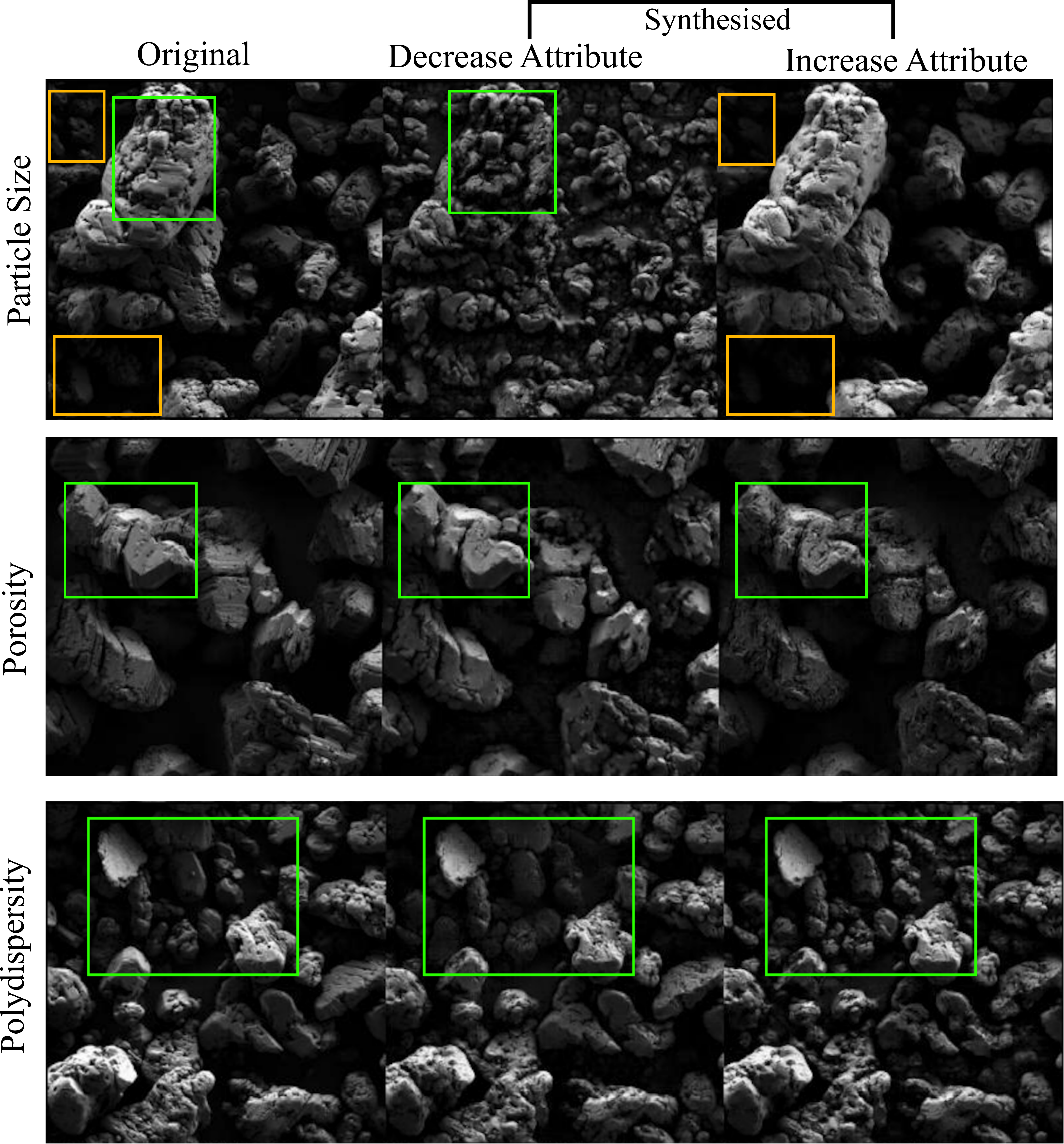}
 \caption{
  Illustration of material attributes controlled SEM image synthesis. The left column is the original SEM image. The middle and right column shows the GAN modified images that alters the corresponding material attributes.
}
\label{fig:attEdit}
\end{figure}

After obtaining these attributes, we train the attGAN~\cite{attGAN} that allows us to selectively edit the salient attributes of a given SEM image..
Since there are 30 lots in total, we essentially only have 30 unique labels (all images from the same lot are given the same attribute values).
Besides the sparsity in labeling information, the other challenges originate from the salience pattern in the images itself.
For example, the porosity of a material is reflected by the presence of small pinprick holes on the surface of the crystals in the SEM image, which only occupies an extremely small number of pixels. Modeling concepts represented by such a minuscule feature can be very challenging.
Despite these obstacles, as illustrated in Figure~\ref{fig:attEdit}, by utilizing the attGAN, the concept-guided synthesis can accurately capture these meaningful material properties.
Such a success not only indicates the accuracy of the estimated material properties by the scientists but also demonstrate the coherency among images from the same lot.

\subsection{Actionable Explanation Pipeline}
\label{sec:forwardbackward}
As illustrated in Figure~\ref{fig:pipeline}, once a synthesis image is generated, we can then feed it into the regression model to predict the respective mechanical properties (e.g., peak stress).
%
The most straightforward way to ascertain the relationship between the material properties and the predicted mechanical properties is to do a simple ``forward'' sensitivity analysis by observing how predicted stress changes as we varying the material properties in the image synthesis process.
To understand the impact of a particular attribute, we can fix all other attributes while varying the value of the attribute of interests.

However, there are some fundamental limitations to this interpretation approach. Firstly, all attribute combinations can be extremely large for simple combinatorial exploration. As a result, we can only meaningfully vary one or two attributes at a time for exploring their impact while fixing all others. Therefore, we cannot effectively study the combined effect of all attributes on the prediction. Also, the forward interpretation cannot directly answer the retrospective question that starting from the model output, e.g., to increase the output peak stress how should the input attribute change?
To answer a question like this, we want to find the necessary changes to the actionable attribute of the input SEM image for obtaining the desired predicted peak stress output.
Let us define the generative editing model as ${G}(\mathbf{I};\mathbf{A})$, where $I$ is the original image and ${\mathbf{A}}=\{a_1,\cdots,a_N\}$ are the material attributes that control the editing.
Given an SEM image $\mathbf{I}$ for which the deep regressor $R$ predicts a peak stress (output) value $p$, we aim to identify the attribute $\mathbf{A}$ such that the edited image $\mathbf{I}'={G}(\mathbf{I};\mathbf{A})$ would lead to a higher/lower peak stress prediction.
Given an image $\mathbf{I}$ with corresponding image attribute vector $\mathbf{A}$ and a target attribute vector $\mathbf{A'}$, which can be solved efficiently using gradient descent.
\label{opt}
\begin{equation}
\begin{aligned}
\min_{\mathbf{A}'} \quad & \|\mathbf{I}-\mathbf{I}(A')\|_p\\
\textrm{s.t.} \quad & p=R(\mathbf{I}(\mathbf{A}'))\\
  & I(\mathbf{A}') = G(\mathbf{I};\mathbf{A}')    \\
\end{aligned}
\end{equation}
where $p$ is the desired output peak stress prediction.

The neural network based regression models make the formulation \ref{opt} non-linear and hard to solve directly. Here, we formulate a relaxed version (as shown below) that can be solved efficiently using gradient descent.

\begin{equation}
\label{opt1}
\begin{aligned}
\min_{A'} \quad & \lambda \cdot loss_{R,p'}(\mathbf{I}(\mathbf{A}'))+  \|\mathbf{I}- \mathbf{I}(\mathbf{A}')\|_p\\
\end{aligned}
\end{equation}

where loss $loss_{R,p'}$ is mean squared error (MSE) loss for predicting image $\mathbf{I}(\mathbf{A}')$ to value $p'$ using the regressor $R$. Since, both regressor $R$ and generator $G$ are differentiable, we can compute the gradient of the objective function via back-propagation and solve the optimization using gradient descent.

%% file: results.tex
\section{Results}
\label{sec:result}

Here we illustrate how we can apply the proposed techniques to help the material scientists obtain insights from the regression model and infer the underlying material science principals.
In this application, it is crucial to understand the relationship between the material's structural features (i.e., features captured by SEM image) and the mechanical properties of the feedstock material.
A deep neural network regression model is trained to predict the peak stress of a material \emph{Lot} from a given SEM image tile.
The regression model is built upon the WideRestNet CNN architecture~\cite{zagoruyko2016wide} and trained on all 30 \emph{Lots}.

\begin{figure}[!htbp]
\centering
  \includegraphics[width=0.99\linewidth]{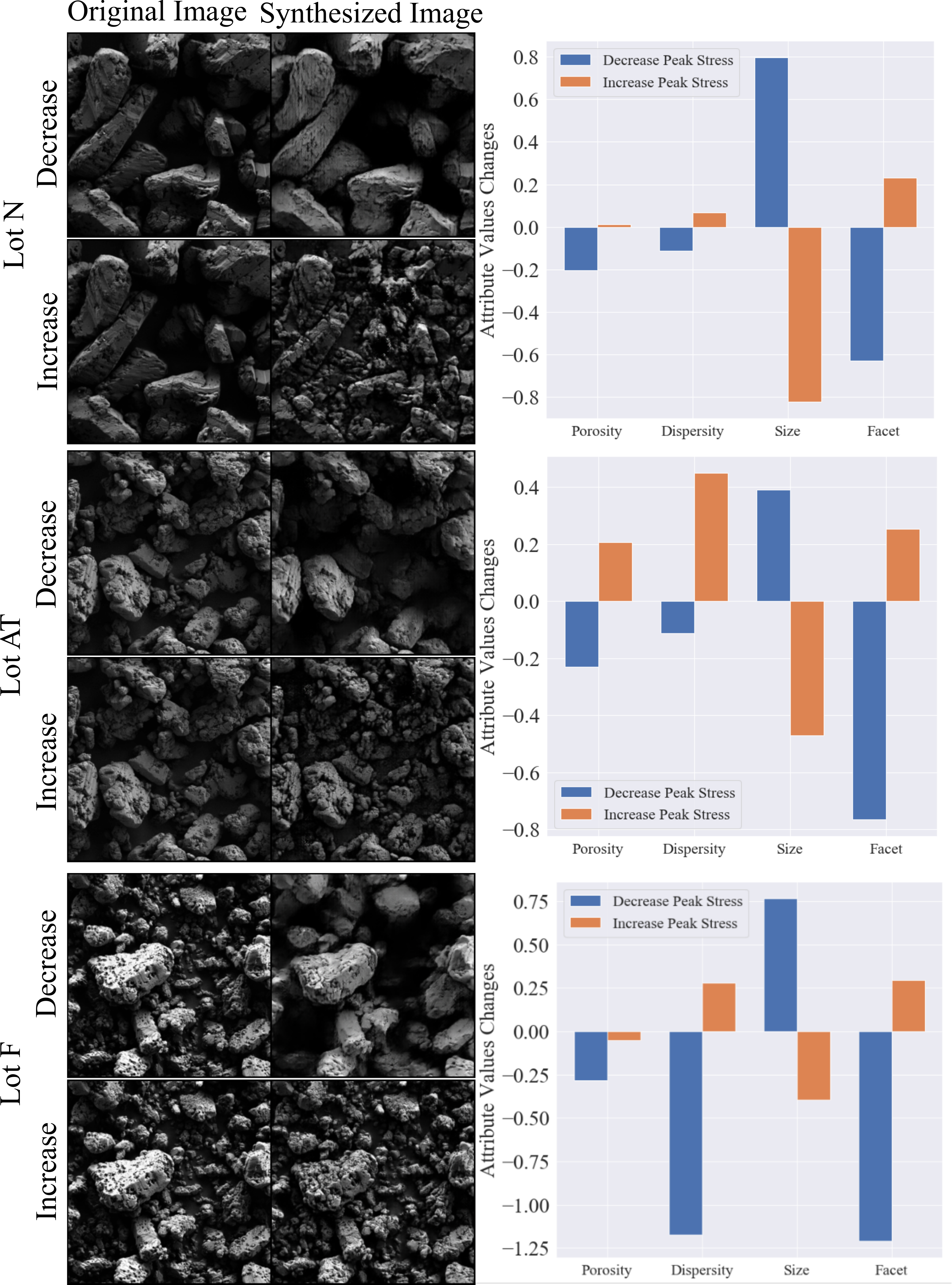}
 \caption{
 The concept-based explanation for single SEM image. The original and edited image tile is shown on the left, and the attribute changes that lead to increase and decrease of the predicted peak stress is illustrated in the plot on the right.
}

\label{fig:backwardPerInstance}
\end{figure}

As shown in Figure~\ref{fig:backwardPerInstance}, we illustrate the explanation results on three SEM images from three different \emph{Lots}, low (N), median (AT), high (F), respectively.
In Figure~\ref{fig:backwardPerInstance}, the original image and modified synthesized image (based on the change in attributes) are shown on the left, whereas the corresponding attribute changes for increasing and decreasing predicted peak stress are ploted on the right.
In the top row (Lot N), we can see in both SEM images (left) and attribute bar-plot (right), that decreasing crystal \emph{size}, while increasing \emph{porosity}, \emph{dispersity}, \emph{facetness} lead to a higher peak stress prediction.
The same pattern can be observed for Lot AT (mid-row).
The bottom row (Lot F) shows a slight deviate for the existing pattern (only difference is in porosity where both changes are negative), however, the small absolute value indicates the change in \emph{porosity} does not really contribute much to the changes in the synthesized image.
One thing to note is that the increase of the \emph{facetness} attributes in the image synthesis process seems to also lead to a marked increase in \emph{dispersity} and a reduction of average \emph{size} (see Section~\ref{sec:imageSynthesis}), so the effect we observe for altering \emph{facetness} is likely also due to the changes in \emph{size} attribute.

%

%